\pdfoutput=1
\documentclass[11pt]{article}

\usepackage[utf8]{inputenc}
\usepackage[T1]{fontenc}
\usepackage{mathptmx}

\usepackage[margin=1in]{geometry}
\usepackage{setspace}
\usepackage{amsmath,amssymb,amsthm}

\usepackage{booktabs}
\usepackage{tabularx}
\usepackage{adjustbox}
\usepackage{multirow}

\usepackage{graphicx}
\usepackage{xcolor}

\usepackage[
    colorlinks=true,
    linkcolor=blue!70!black,
    citecolor=blue!70!black,
    urlcolor=blue!70!black,
    bookmarksnumbered=true
]{hyperref}

\usepackage{enumitem}
\setlist{nosep, leftmargin=*}

\usepackage{microtype}

\usepackage{fancyhdr}
\usepackage{titlesec}
\titleformat{\section}{\large\bfseries}{\thesection}{1em}{}
\titleformat{\subsection}{\normalsize\bfseries}{\thesubsection}{1em}{}
\titleformat{\subsubsection}{\normalsize\itshape}{\thesubsubsection}{1em}{}

\usepackage{abstract}

\usepackage{float}

\newtheorem{definition}{Definition}
\newtheorem{hypothesis}{Hypothesis}

\begin{document}

% === Title Block ===
\begin{center}
{\Large\bfseries AI Integrity: A New Paradigm for Verifiable AI Governance}\\[0.8em]
{\small\textsc{A Working Paper}}\\[1.2em]
{\large\textbf{Seulki Lee}}\\[0.3em]
{\small AI Integrity Organization (AIO), Geneva, Switzerland}\\
{\small \href{mailto:2sk@aioq.org}{2sk@aioq.org} \quad|\quad \href{https://aioq.org}{aioq.org}}\\[0.8em]
{\small April 2026}\\[0.5em]
{\footnotesize CC BY 4.0 \quad|\quad AIO Working Paper}\\
{\footnotesize Companion to: S.~Lee (2026b), S.~Lee (2026c)}
\end{center}

\vspace{1em}

% === Abstract ===
\begin{abstract}
\noindent
AI systems increasingly shape high-stakes decisions in healthcare, law, defense, and education, yet existing governance paradigms---AI Ethics, AI Safety, and AI Alignment---share a common limitation: they evaluate outcomes rather than verifying the reasoning process itself. This paper introduces AI Integrity, a concept defined as a state in which the Authority Stack of an AI system---its layered hierarchy of values, epistemological standards, source preferences, and data selection criteria---is protected from corruption, contamination, manipulation, and bias, and maintained in a verifiable manner. We distinguish AI Integrity from the three existing paradigms, define the Authority Stack as a 4-layer cascade model (Normative, Epistemic, Source, and Data Authority) grounded in established academic frameworks---Schwartz Basic Human Values for normative authority, Walton argumentation schemes with GRADE/CEBM hierarchies for epistemic authority, and Source Credibility Theory for source authority---characterize the distinction between legitimate cascading and Authority Pollution, and identify Integrity Hallucination as the central measurable threat to value consistency. We further specify the PRISM (Profile-based Reasoning Integrity Stack Measurement) framework as the operational methodology, defining six core metrics and a phased research roadmap. Unlike normative frameworks that prescribe which values are correct, AI Integrity is a procedural concept: it requires that the path from evidence to conclusion be transparent and auditable, regardless of which values a system holds. This shift from outcome evaluation to process verification opens a new approach to AI governance grounded in empirical measurement.
\end{abstract}


\vspace{1em}
\hrule
\vspace{1em}

% ==========================================
\section{Introduction}
% ==========================================

When an AI system recommends a medical treatment, denies a loan application, or flags a person as a security risk, a fundamental question arises: can we verify the reasoning that produced that output? Not merely whether the output was statistically correct, but whether the values, evidence, and sources that led to it were handled transparently, consistently, and in accordance with the system's stated principles?

Current AI governance frameworks cannot answer this question. They evaluate outputs---whether behavior is harmful, whether preferences are satisfied---but do not examine the reasoning process itself. A system can produce nominally acceptable outputs while systematically applying incoherent value hierarchies, selectively weighting evidence, or demoting inconvenient sources. None of these process-level failures are captured by outcome-based evaluation.

This paper introduces AI Integrity as a distinct concept in AI governance, defined by a simple demand: that the reasoning process of an AI system be transparent, consistent, and auditable. AI Integrity does not prescribe which values are correct. It requires that whatever value hierarchy a system holds be empirically measurable and that the path from evidence to conclusion be traceable.

Our contributions are fivefold:

\begin{enumerate}
\item We define AI Integrity as a concept distinct from AI Ethics, AI Safety, and AI Alignment, clarifying the structural gap these paradigms leave unfilled.
\item We propose the Authority Stack---a 4-layer cascade model decomposing AI decision-making into Normative, Epistemic, Source, and Data Authority---as the structural object that AI Integrity requires to be protected and verifiable.
\item We ground each layer in established academic frameworks: Schwartz Basic Human Values \cite{schwartz1992,schwartz2012} for Layer~4, Walton argumentation schemes \cite{walton2008} with GRADE/CEBM evidence hierarchies for Layer~3, and Source Credibility Theory \cite{hovland1953,pornpitakpan2004} for Layer~2.
\item We define Authority Pollution and Integrity Hallucination as the two primary threats to AI Integrity, and characterize the conditions under which inter-layer influence constitutes a violation of reasoning transparency.
\item We specify the PRISM framework---including the Enhanced Cascade Mapping Hypothesis, six core metrics, and a phased research roadmap---as a concrete, reproducible measurement program.
\end{enumerate}

% ==========================================
\section{Existing Paradigms and the Gap They Leave}
% ==========================================

\subsection{Three Paradigms of AI Governance}

\textbf{AI Ethics} asks `Is this AI morally right?' \cite{floridi2019,jobin2019}. Ethics establishes normative aspirations and provides evaluative standards for AI outputs. Its limitation is prescriptive: it adjudicates which values are correct rather than verifying that the system's actual reasoning reflects any consistent value structure. A system can produce Ethics-compliant outputs while operating on an opaque or incoherent value hierarchy.

\textbf{AI Safety} asks `Is this AI system secure from harm?' \cite{amodei2016}. Safety addresses external robustness---preventing adversarial manipulation, distributional shift failures, and reward hacking. It does not examine how the system internally selects and weights evidence. A system can be safe from external attack while systematically applying biased epistemic standards.

\textbf{AI Alignment} asks `Does this AI do what humans want?' \cite{bai2022,ouyang2022}. Alignment requires the premise that a target preference set has been agreed upon---a problematic assumption when deploying across diverse cultural and institutional contexts \cite{gabriel2020}. Even successful alignment does not reveal the mechanism by which the system reached aligned outputs: whether through a coherent value structure or through superficial pattern matching.

\textbf{AI Integrity} asks a structurally different question: `Can we verify what values, evidence, sources, and data this AI used to reach its conclusion?' It does not prescribe which values are correct. It requires that the entire reasoning process be transparent, measurable, and auditable. This shifts the focus from outcome evaluation to process verification.

\begin{table}[H]
\centering
\caption{Four AI governance paradigms compared}
\label{tab:paradigms}
\small
\begin{tabularx}{\textwidth}{lXXXX}
\toprule
\textbf{Paradigm} & \textbf{Core Question} & \textbf{Focus} & \textbf{Key Limitation} & \textbf{Evaluation Mode} \\
\midrule
Ethics & Morally right? & Value judgments & Whose values? & Normative \\
Safety & Secure from harm? & Attack prevention & Ignores internal bias & Defensive \\
Alignment & Does what humans want? & Preference matching & Whose preferences? & Outcome-based \\
\textbf{Integrity} & \textbf{Reasoning verifiable?} & \textbf{Stack transparency} & \textbf{Requires measurement} & \textbf{Process-based} \\
\bottomrule
\end{tabularx}
\end{table}

\subsection{`Integrity' in Prior AI Literature}

`Integrity' has appeared in existing AI literature in three narrower senses. The EU AI Act (2024) refers to `system integrity' primarily from a cybersecurity perspective---protection against unauthorized access and data manipulation. NIST AI RMF (2023) uses `data integrity' as accuracy and completeness of training data. IEEE Ethically Aligned Design (2019) invokes integrity in the context of professional ethics for AI practitioners. None of these usages addresses the full chain of values-evidence-reasoning. Our definition is broader and more operationally specific: AI Integrity encompasses the entire Authority Stack maintaining verifiable transparency, not merely the security or accuracy of one of its components.

\subsection{Why Process Verification Is Not Yet Addressed}

The three paradigms share a common architecture: they specify properties that AI outputs should satisfy (moral rightness, harmlessness, preference alignment) and evaluate systems by testing whether outputs satisfy those properties. This architecture makes two implicit assumptions: that output properties can be fully specified in advance, and that output-level evaluation is sufficient to verify reasoning quality.

Both assumptions are questionable. Specifying output properties in advance requires anticipating every context in which the system will operate---an intractable combinatorial problem for systems deployed across diverse professional domains. And output-level evaluation cannot distinguish between a system that reached the right answer through coherent reasoning and one that reached it through pattern matching on surface features while applying an incoherent or biased internal process. Process verification requires examining the reasoning structure itself, not merely its outputs.

% ==========================================
\section{AI Integrity: Definition}
% ==========================================

\begin{definition}[AI Integrity]
A state in which the Authority Stack of an AI system---its layered hierarchy of values, epistemological standards, source preferences, and data selection criteria---is protected from corruption, contamination, manipulation, and bias, and maintained in a verifiable manner.
\end{definition}

AI Integrity is not about whether an AI gives the `correct' answer. It is about whether the path from evidence to conclusion is transparent and auditable. This framework does not adjudicate truth; it standardizes verification of the path to truth.

Three properties define the concept.

\textbf{Process-oriented, not outcome-prescriptive.} AI Integrity does not dictate which values are correct. It requires that the value selection process be measurable and auditable. A system that consistently applies a coherent value hierarchy satisfies the integrity criterion, regardless of which specific values it holds. What violates AI Integrity is not the presence of particular values but the opacity or inconsistency of the value selection process itself.

\textbf{Multi-layered.} AI Integrity requires examination of the full Authority Stack---not just the values a system claims to hold (Layer~4), but also what evidence it counts as valid (Layer~3), which sources it trusts (Layer~2), and what data it selects (Layer~1). Single-layer evaluation---assessing values alone---produces incomplete integrity assessments and can miss systematic distortions in the layers beneath.

\textbf{Verifiable.} AI Integrity demands empirical measurement, not self-report or stated principles. A system's integrity profile must be derivable from observable behavioral patterns across systematically varied conditions. Stated commitments to transparency do not constitute AI Integrity; measured behavioral consistency does.

Three operational axes implement this definition in practice.

\textbf{Value Consistency:} The ability to maintain a coherent value hierarchy across contextual variations---across different professional domains, levels of severity, and decision timeframes.

\textbf{Judgmental Accountability:} All AI judgments should be grounded in a transparent Authority Stack that is externally auditable.

\textbf{Agency Protection:} AI should assist rather than replace human autonomous judgment, without covertly steering toward particular conclusions through selective evidence presentation or opaque source weighting.

% ==========================================
\section{The Authority Stack: A 4-Layer Model}
% ==========================================

\subsection{Layer Architecture}

The structural object that AI Integrity requires to be protected is the Authority Stack: the complete hierarchy of normative and epistemic commitments that govern AI reasoning.

\begin{table}[H]
\centering
\caption{Authority Stack layer architecture}
\label{tab:authority-stack}
\small
\begin{tabularx}{\textwidth}{clXXX}
\toprule
\textbf{Layer} & \textbf{Name} & \textbf{Governing Question} & \textbf{Content} & \textbf{Theoretical Basis} \\
\midrule
L4 & Normative Authority & What values guide the decision? & Value hierarchy---the ordering of competing values when they conflict & Schwartz Basic Human Values \cite{schwartz1992,schwartz2012} \\
L3 & Epistemic Authority & What evidence types are valid? & Evidence standards---which types of evidence are treated as decisive & Walton Argumentation Schemes \cite{walton2008} + GRADE/CEBM \\
L2 & Source Authority & Which sources are trusted? & Source credibility judgments---which institutions or actors are deferred to & Source Credibility Theory \cite{hovland1953} \\
L1 & Data Authority & What data is selected? & Data selection---what information enters the reasoning process and what is excluded & Derived variable: L1 = $f$(L4, L3, L2) \\
\bottomrule
\end{tabularx}
\end{table}

\subsection{Theoretical Grounding of Each Layer}

Each of the three independently measurable layers is grounded in an established academic framework that provides both a classification taxonomy and a theoretical rationale for the categories employed.

\textbf{Layer 4 --- Schwartz Basic Human Values.} The normative authority layer employs Schwartz's theory of basic human values \cite{schwartz1992,schwartz2012}, which identifies 10 basic values (Universalism, Benevolence, Conformity, Tradition, Security, Power, Achievement, Hedonism, Stimulation, Self-Direction) organized in a circular motivational continuum under four higher-order categories (Self-Transcendence, Conservation, Self-Enhancement, Openness to Change). This framework was selected for three reasons. First, it is the most extensively cross-culturally validated value theory available, tested across over 80 countries \cite{schwartz2012}. Second, the circular structure specifies motivational compatibilities and conflicts between values---adjacent values on the circle share motivational goals while opposing values conflict---providing a theoretically grounded expectation for which value trade-offs should be most difficult. Third, the 10-value level yields $C(10,2) = 45$ pairwise comparisons, a tractable number for forced-choice benchmark design. While the refined theory \cite{schwartz2012r} distinguishes 19 sub-values, the 10-value level maintains measurement parsimony without sacrificing theoretical structure.

\textbf{Layer 3 --- Walton Argumentation Schemes + GRADE/CEBM.} The epistemic authority layer draws on two complementary frameworks. Walton, Reed, and Macagno's \cite{walton2008} argumentation schemes provide a domain-agnostic typology of reasoning patterns---argument from expert opinion, argument from evidence to hypothesis, argument from analogy, and others---that classify the logical structure of evidence claims independent of subject matter. This is supplemented by the GRADE and CEBM evidence hierarchies, which rank evidence types by methodological strength (from systematic reviews at the top to expert opinion at the bottom). The combination yields 10 evidence types that span the full range from systematic synthesis to popular consensus, each distinguishable in a forced-choice scenario without requiring specialized domain knowledge.

\textbf{Layer 2 --- Source Credibility Theory.} The source authority layer integrates Walton's source-related argumentation schemes with the three credibility dimensions identified in the foundational work of Hovland, Janis, and Kelley \cite{hovland1953} and synthesized in Pornpitakpan's \cite{pornpitakpan2004} comprehensive review: competence (perceived expertise), trustworthiness (perceived objectivity and reliability), and goodwill (perceived concern for the audience's interests). These three dimensions yield 10 source types ranging from international organizations (high competence + trustworthiness) through direct stakeholders (high experiential standing) to anonymous crowdsourced information (low traceability). The typology captures the full spectrum of sources that AI systems may differentially weight in reasoning.

\textbf{Layer 1 --- Derived Variable.} Layer~1 (Data Authority) is not independently measured. It is derived as a function of the three upper-layer profiles: L1 = $f$(L4, L3, L2). The theoretical claim is that a system's data selection patterns are determined by the interaction of its value commitments (which topics matter), epistemic standards (which evidence types are sought), and source preferences (which providers are consulted). Whether this derivation holds empirically is a testable hypothesis---one that the PRISM framework is designed to evaluate through the Cascade Consistency Index.

\subsection{The Cascade Structure}

The four layers operate as a cascade: higher layers constrain lower layers.

\begin{verbatim}
L4: Normative Authority -- What values guide the decision?
    Basis: Schwartz Basic Human Values (2012)
         | constrains
         v
L3: Epistemic Authority -- What evidence types are considered valid?
    Basis: Walton Argumentation Schemes + GRADE/CEBM
         | constrains
         v
L2: Source Authority -- Which sources are trusted?
    Basis: Source Credibility Theory (Hovland et al., 1953)
         | determines
         v
L1: Data Authority -- What data is selected or excluded?
    Derived variable: L1 = f(L4, L3, L2)
\end{verbatim}

A critical clarification is required. The top-down cascade is a measurement-oriented analytical framework, not a claim about the causal architecture of language models. In practice, pre-training data distributions (L1) shape the value orientations that emerge during fine-tuning (L4), and RLHF and Constitutional AI processes operate on all layers simultaneously rather than sequentially. We adopt the top-down cascade as an analytical structure because it provides a tractable framework for independent layer measurement and inter-layer consistency assessment. The key claim is not `L4 causally determines L3,' but rather: a coherent AI system should exhibit consistent patterns across layers, and the degree of this consistency is empirically measurable. Whether the cascade accurately models internal model dynamics is itself an empirical question---one that the Cascade Consistency Index (CCI) is designed to test. If CCI is consistently near 0.5 (chance level), the cascade model itself requires revision. The hypothesis is thus falsifiable by design.

\subsection{Legitimate Cascading vs. Authority Pollution}

Inter-layer influence is not inherently a problem. Authority Pollution is a different phenomenon: inter-layer influence that is opaque, inconsistent, or that distorts facts in ways not justified by the value system.

\begin{table}[H]
\centering
\caption{Legitimate cascading vs.\ Authority Pollution}
\label{tab:pollution}
\small
\begin{tabularx}{\textwidth}{lXX}
\toprule
 & \textbf{Legitimate Cascading} & \textbf{Authority Pollution} \\
\midrule
\textbf{Mechanism} & Values calibrate judgment standards & Values distort or suppress facts \\
\textbf{Example} & `Patient safety first' (L4) $\rightarrow$ raise medical evidence bar (L3) & `Inclusivity first' (L4) $\rightarrow$ ignore demographic data entirely (L1) \\
\textbf{Character} & Consistent and predictable influence & Opaque and selective influence \\
\textbf{Traceability} & Explicable from the value system & Not explicable; facts were omitted without declared reason \\
\bottomrule
\end{tabularx}
\end{table}

Three documented cases illustrate the range of Authority Pollution types:

\textbf{Case 1 --- L4$\rightarrow$L1 Pollution (Documented):} Google's Gemini generated historically inaccurate images in 2024, depicting Nazi-era German soldiers and the American Founding Fathers with racial diversity inconsistent with historical record. An L4 normative directive (inclusivity) contaminated L1 data selection (historical facts), overriding established evidence to satisfy a value priority \cite{grant2024}.

\textbf{Case 2 --- L4$\rightarrow$L3 Pollution (Illustrative Construct):} A healthcare AI that generates reassuring but clinically inaccurate responses when presented with symptoms of serious conditions---effectively prioritizing Benevolence (patient comfort, L4) over evidence accuracy (L3) \cite{thirunavukarasu2023}.

\textbf{Case 3 --- L2$\rightarrow$L1 Pollution (Illustrative Pattern):} When an AI system systematically ranks institutional sources above community-sourced information regardless of informational content or contextual relevance, source preferences determine data selection independent of data quality.

% ==========================================
\section{Integrity Hallucination}
% ==========================================

A distinct threat to AI Integrity is Integrity Hallucination: the phenomenon in which an AI system provides different value judgments for structurally identical scenarios, generating probabilistically plausible ethical statements while lacking a coherent underlying value structure.

\begin{table}[H]
\centering
\caption{Three mechanisms of Integrity Hallucination}
\label{tab:integrity-hallucination}
\small
\begin{tabularx}{\textwidth}{lXX}
\toprule
\textbf{Mechanism} & \textbf{Symptom} & \textbf{Governance Implication} \\
\midrule
Stochastic variation & Low test-retest reliability; stable aggregate win-rates & Benign; addressed by temperature calibration \\
Framing sensitivity & Choice reversal across narrative perspectives; unstable under prompt variation & Moderate; value hierarchy exists but shallowly held \\
Structural incoherence & No consistent win-rate pattern; responses reflect pattern-matching, not principled application & Severe; system lacks any stable value structure to audit \\
\bottomrule
\end{tabularx}
\end{table}

Preliminary empirical evidence from forced-choice value judgment experiments across 10 AI models (S.~Lee, 2026b) confirms that Integrity Hallucination is observable across all tested models, with consistency rates varying from 62\% to 94\% across value pairs. The PRISM framework operationalizes the decomposition of these three mechanisms through two complementary metrics---Test-Retest Reliability (TRR) and Scenario Replication Score (SRS)---whose joint analysis enables differential diagnosis: high TRR with low SRS indicates framing sensitivity, low TRR with high SRS indicates stochastic noise, and low scores on both indicate structural incoherence. This diagnostic capacity transforms Integrity Hallucination from a qualitative concern into a quantitatively measurable phenomenon.

% ==========================================
\section{The PRISM Framework}
% ==========================================

\subsection{Enhanced Cascade Mapping Hypothesis}

The central empirical hypothesis underlying the PRISM (Profile-based Reasoning Integrity Stack Measurement) framework is the Enhanced Cascade Mapping Hypothesis:

\begin{hypothesis}[Enhanced Cascade Mapping]
When Layer~4 (value priorities), Layer~3 (evidence-type priorities), and Layer~2 (source-type priorities) are each measured independently through forced-choice benchmarks, Layer~1 (data selection patterns) can be derived as a function of the three upper-layer profiles:
\begin{equation}
\text{L1}_{\text{predicted}} = f(\text{L4}_{\text{profile}},\; \text{L3}_{\text{profile}},\; \text{L2}_{\text{profile}})
\end{equation}
Furthermore, the model's final response to any given scenario becomes predictable from its PRISM profile---enabling pre-deployment behavioral forecasting.
\end{hypothesis}

Rather than deriving L3/L2 patterns from L4 analytically and assuming their consistency, the Enhanced Hypothesis requires independent measurement of each layer---making cascade coherence an empirical question rather than a theoretical assumption. The hypothesis is falsifiable by design: if CCI is consistently near chance level (0.5), the cascade model itself requires revision.

\subsection{Operational Methodology}

The PRISM measurement program proceeds in five steps:

\begin{enumerate}
\item Independently measure three observable layers (L4, L3, L2) using forced-choice benchmarks grounded in the academic frameworks specified in Section~4.2.
\item For each measured layer, generate predictions for other layers using cascade mapping rules derived from the theoretical relationships between value commitments and epistemic/source standards.
\item Compare predictions against independent measurements to compute CCI. Compute SRS and TRR to assess measurement reliability.
\item Derive L1 from the three measured profiles and validate through targeted free-form response prediction experiments.
\item Test ASPA---whether the full PRISM profile can predict free-form model responses to novel scenarios at rates sufficient for pre-deployment behavioral forecasting.
\end{enumerate}

\subsection{Core Metrics}

\begin{table}[H]
\centering
\caption{PRISM core metrics}
\label{tab:metrics}
\footnotesize
\begin{tabularx}{\textwidth}{llXlX}
\toprule
\textbf{Metric} & \textbf{Full Name} & \textbf{Measures} & \textbf{Range} & \textbf{Interpretation} \\
\midrule
VE & Value Entropy & Dispersion of value judgments across all pairings & 0 to $\log_2(n)$ & Low = strong hierarchy; High = dispersed preferences \\
SRS & Scenario Replication Score & Choice stability across 3 independent scenario instantiations & 0.33--1.0 & High = condition-level generalizability \\
TRR & Test-Retest Reliability & Choice stability on identical scenarios presented twice & 0--1.0 & High = stochastic stability \\
CCI & Cascade Consistency Index & Alignment between independently measured L3/L2 and L4-predicted L3/L2 & 0--1.0 & $>$0.5 = cascade operating \\
ASPA & Auth.\ Stack Predictive Accuracy & Accuracy of PRISM profile in predicting free-form responses & 0--1.0 & Ultimate validation \\
PCS & Perspective Consistency Score & Choice stability across 5 stakeholder-perspective re-framings & 0.33--1.0 & High = framing-robust judgment \\
\bottomrule
\end{tabularx}
\end{table}

SRS and TRR together enable decomposition of Integrity Hallucination into three distinct mechanisms:

\begin{table}[H]
\centering
\caption{Integrity Hallucination decomposition via TRR $\times$ SRS}
\label{tab:ih-decomposition}
\small
\begin{tabularx}{\textwidth}{ccXX}
\toprule
\textbf{TRR} & \textbf{SRS} & \textbf{Diagnosis} & \textbf{Governance Implication} \\
\midrule
High & High & Genuine value hierarchy & Reliable for deployment; PRISM profile is predictive \\
High & Low & Framing sensitivity & Value hierarchy exists but surface-level; prompt design matters \\
Low & High & Stochastic noise & Hierarchy may exist; aggregate statistics reliable but individual responses variable \\
Low & Low & \textbf{Structural incoherence} & No coherent value hierarchy; unpredictable in high-stakes deployment \\
\bottomrule
\end{tabularx}
\end{table}

\subsection{Research Roadmap}

\begin{table}[H]
\centering
\caption{PRISM phased research program}
\label{tab:roadmap}
\small
\begin{tabularx}{\textwidth}{clXl}
\toprule
\textbf{Phase} & \textbf{Period} & \textbf{Target \& Success Criterion} & \textbf{Key Output} \\
\midrule
Phase~1 & 2026 Q2 & L4 Refinement: fixed scenarios, SRS + TRR. Success: TRR $\geq$ 0.75 across $\geq$ 80\% of value pairs & IH decomposition \\
Phase~2 & 2026 Q3 & L3 Benchmark: evidence-type priorities, first CCI. Success: CCI $>$ 0.60 in $\geq$ 3 of 7 domains & Epistemic Authority in AI \\
Phase~3 & 2026 Q4 & L2 Benchmark: source-type priorities, integrated CCI. Success: L2 CCI $\geq$ L3 CCI in $\geq$ 4 of 7 domains & Source Authority Profiles \\
Phase~4 & 2027 Q1 & Integration: L1 derivation, ASPA. Success: ASPA $>$ 0.65 on held-out scenarios & Predicting AI Judgment \\
\bottomrule
\end{tabularx}
\end{table}

The PRISM framework is designed to be compatible with neuro-symbolic governance architectures such as GRACE \cite{jahn2026}. While GRACE provides the governance mechanism for enforcing normative constraints, PRISM provides the measurement substrate: PRISM profiles could inform the normative specifications that architectures like GRACE enforce, creating a measurement-to-governance pipeline.

\subsection{Relationship to This Paper}

The relationship between AI Integrity (this paper) and PRISM is analogous to the relationship between a governance concept and its auditing methodology. AI Integrity defines what `verifiable reasoning' means and why it is necessary. PRISM specifies how to measure whether a given system achieves it. A companion empirical paper (S.~Lee, 2026b) provides preliminary evidence from L4 measurement, and a companion risk signal paper (S.~Lee, 2026c) demonstrates the application of hierarchy-based measurement to AI safety governance.

% ==========================================
\section{Limitations and Open Questions}
% ==========================================

\subsection{Acknowledged Limitations}

\textbf{Procedural neutrality and its limits.} AI Integrity is explicitly a procedural concept: it requires transparency and consistency of the reasoning process without prescribing which values are correct. A system that consistently applies a harmful value hierarchy with perfect internal coherence satisfies the AI Integrity criterion. AI Integrity is therefore a necessary but not sufficient condition for trustworthy AI.

\textbf{The cascade as analytical abstraction.} The Authority Stack is presented as a top-down cascade, but this is an analytical framework, not a causal claim. Language models are not architecturally organized into four discrete layers. The degree to which the cascade model captures meaningful structure in AI reasoning is itself an empirical question.

\textbf{Distinguishing pollution from legitimate complexity.} The distinction between Authority Pollution and legitimate contextual variation is not always clear-cut.

\textbf{Scope of the concept.} AI Integrity as defined here applies to AI systems that produce reasoning-grounded outputs in professional contexts. It is most relevant for large language models and other systems that generate rationale-bearing outputs in high-stakes domains.

\textbf{Temporal scope.} The current framework measures AI value profiles at a specific point in time. Model updates, fine-tuning iterations, and evolving training data may shift profiles between measurements. AI Integrity assessment should therefore be understood as periodic audit rather than permanent certification.

\textbf{Walton scheme applicability to AI.} Walton's argumentation schemes were designed for human reasoning. Whether AI `evidence preferences' operate under the same structural logic as human argumentation requires empirical validation through the PRISM L3 benchmark.

\textbf{Forced-choice binary simplification.} Real-world AI decisions involve multi-factor weighing, not binary choices. The forced-choice protocol reveals priority orderings but not weighting magnitudes.

\subsection{Open Questions}

\begin{enumerate}
\item Does Integrity Hallucination occur independently at each layer, or is it a whole-stack phenomenon?
\item Are entropy patterns correlated across layers? Do models with strong L4 hierarchies also show strong L3/L2 hierarchies?
\item What ASPA threshold qualifies as `understanding' a model's Authority Stack? At what predictive accuracy can pre-deployment behavioral forecasting be meaningful?
\item How do model updates affect Authority Stack stability over time? Is a PRISM profile a stable property or a snapshot?
\item Can mechanistic interpretability methods identify the internal structures that produce specific PRISM profile patterns?
\end{enumerate}

% ==========================================
\section{Conclusion}
% ==========================================

This paper has introduced AI Integrity as a distinct concept in AI governance---one that fills a structural gap left by Ethics, Safety, and Alignment: the absence of any framework for verifying the reasoning process itself rather than its outputs.

The Authority Stack model provides the structural object that AI Integrity requires to be protected: a four-layer hierarchy in which values constrain evidence standards, which constrain source preferences, which determine data selection. Each of the three independently measurable layers is grounded in an established academic framework---Schwartz Basic Human Values for normative authority, Walton argumentation schemes with GRADE/CEBM hierarchies for epistemic authority, and Source Credibility Theory for source authority---providing both theoretical justification for the layer decomposition and a concrete taxonomy for measurement.

Integrity Hallucination identifies the condition in which no coherent layer structure exists at all. The Enhanced Cascade Mapping Hypothesis, operationalized through the PRISM framework with six core metrics and a phased research roadmap, provides a falsifiable empirical program for testing whether the cascade model captures meaningful structure in AI reasoning.

AI Integrity does not require consensus on which values are correct. By requiring that reasoning paths be measurable and auditable, it enables diverse stakeholders to make informed choices about which AI systems align with their institutional values---and to verify that alignment empirically rather than taking it on trust.

\bigskip
\noindent\textbf{Declarations.} The author is founder and CEO of AI Integrity Organization (AIO), a Swiss-registered nonprofit (UID: CHE-469.997.903). This research received no external funding.

\medskip
\noindent\textbf{Data Availability.} Published results for public models and supporting papers are available at aioq.org. Benchmark evaluations are conducted by AIO; methodology details are described in the companion papers S.~Lee (2026b) and S.~Lee (2026c).

% ==========================================
% References
% ==========================================


\begin{thebibliography}{99}

\bibitem{amodei2016}
Amodei, D., et al.\ (2016).
Concrete problems in AI safety.
\textit{arXiv:1606.06565}.

\bibitem{bai2022}
Bai, Y., et al.\ (2022).
Training a helpful and harmless assistant with RLHF.
\textit{arXiv:2204.05862}.

\bibitem{euaiact2024}
European Parliament. (2024).
Regulation (EU) 2024/1689 (AI Act).

\bibitem{floridi2019}
Floridi, L., \& Cowls, J. (2019).
A unified framework of five principles for AI in society.
\textit{Harvard Data Science Review}, 1(1).

\bibitem{gabriel2020}
Gabriel, I. (2020).
Artificial intelligence, values, and alignment.
\textit{Minds and Machines}, 30(3), 411--437.

\bibitem{grant2024}
Grant, N. (2024).
Google pauses Gemini AI image generator after historical inaccuracies.
\textit{The New York Times}, Feb.\ 22.

\bibitem{hovland1953}
Hovland, C.~I., Janis, I.~L., \& Kelley, H.~H. (1953).
\textit{Communication and Persuasion}. Yale University Press.

\bibitem{ieee2019}
IEEE. (2019).
\textit{Ethically Aligned Design}, 1st ed.

\bibitem{jahn2026}
Jahn, F., et al.\ (2026).
Breaking up with normatively monolithic agency with GRACE.
\textit{arXiv:2601.10520}.

\bibitem{jobin2019}
Jobin, A., Ienca, M., \& Vayena, E. (2019).
The global landscape of AI ethics guidelines.
\textit{Nature Machine Intelligence}, 1(9), 389--399.

\bibitem{lee2026b}
Lee, S. (2026b).
Measuring AI value priorities: Empirical analysis of forced-choice responses across AI models.
\textit{Preprint}.

\bibitem{lee2026c}
Lee, S. (2026c).
PRISM Risk Signal Framework: Hierarchy-based red lines for AI behavioral risk.
\textit{Preprint}.

\bibitem{nist2023}
NIST. (2023).
AI Risk Management Framework (AI RMF 1.0). NIST AI 100-1.

\bibitem{ouyang2022}
Ouyang, L., et al.\ (2022).
Training language models to follow instructions with human feedback.
\textit{NeurIPS 2022}.

\bibitem{pornpitakpan2004}
Pornpitakpan, C. (2004).
The persuasiveness of source credibility: A critical review of five decades' evidence.
\textit{Journal of Applied Social Psychology}, 34(2), 243--281.

\bibitem{schwartz1992}
Schwartz, S.~H. (1992).
Universals in the content and structure of values.
\textit{Advances in Experimental Social Psychology}, 25, 1--65.

\bibitem{schwartz2012}
Schwartz, S.~H. (2012).
An overview of the Schwartz theory of basic values.
\textit{Online Readings in Psychology and Culture}, 2(1).

\bibitem{schwartz2012r}
Schwartz, S.~H., et al.\ (2012).
Refining the theory of basic individual values.
\textit{Journal of Personality and Social Psychology}, 103(4), 663--688.

\bibitem{thirunavukarasu2023}
Thirunavukarasu, A.~J., et al.\ (2023).
Large language models in medicine.
\textit{Nature Medicine}, 29(8), 1930--1940.

\bibitem{walton2008}
Walton, D., Reed, C., \& Macagno, F. (2008).
\textit{Argumentation Schemes}. Cambridge University Press.

\end{thebibliography}
\end{document}